\documentclass[11pt,a4paper]{article}
\usepackage[hyperref]{emnlp2018}
\aclfinalcopy

\usepackage{times}
\usepackage{latexsym}

\usepackage{url}
\usepackage{pslatex}

\usepackage{amsmath}
\usepackage{caption}
\usepackage{xcolor}
\usepackage{graphicx}
\usepackage{multirow}
\usepackage{pgfplots}
\usepackage{subcaption}
\usepackage{tabularx}
\usepackage{booktabs}
\usepackage{xspace}

\newcommand\ie{\emph{i.e.}}
\newcommand\eg{\emph{e.g.}}
\newcommand\cf{\emph{cf.}\ }

\newcommand{\Table}[1]{Table~\ref{#1}}

\newcommand{\Figure}[1]{Figure~\ref{#1}}

\newcommand{\Section}[1]{Section~\ref{#1}}

\newcommand{\sofb}{\textit{second-order false-belief}\,}
\newcommand{\fb}{\textit{false-belief}\xspace}
\newcommand{\tb}{\textit{true-belief}\xspace}

\newcommand{\tom}{\textit{ToM}\xspace}
\newcommand{\tomeasy}{\textit{ToM-easy}\xspace}

\usepackage{graphicx}

\title{Evaluating Theory of Mind in Question Answering \vspace{25pt} }

\author{
  Aida Nematzadeh\\
  DeepMind \\
  {\tt nematzadeh@google.com} 
  \\\And
  Kaylee Burns\\
  UC Berkeley\\
  {\tt kayleeburns@berkeley.edu}
  \And
  Erin Grant \\
  UC Berkeley\\
  {\tt eringrant@berkeley.edu}
  \AND
  Alison Gopnik \\
  UC Berkeley\\
  {\tt gopnik@berkeley.edu} \\
  \\\And 
  Thomas L.\ Griffiths \\
  Princeton University\\
  {\tt tomg@princeton.edu}\\
  }

\begin{document}
\maketitle

\begin{abstract}

We propose a new dataset for evaluating question answering models with respect to their capacity to reason about beliefs. Our tasks are inspired by theory-of-mind experiments that examine whether children are able to reason about the beliefs of others, in particular when those beliefs differ from reality. We evaluate a number of recent neural models with memory augmentation. We find that all fail on our tasks, which require keeping track of inconsistent states of the world; moreover, the models' accuracy decreases notably when random sentences are introduced to the tasks at test.\footnote{
 Code to generate dataset and replicate results is available at \href{https://github.com/kayburns/tom-qa-dataset}{\texttt{github.com/kayburns/tom-qa-dataset}}.
} 

\noindent
\end{abstract}

\section{Reasoning About Beliefs}

Possessing a capacity similar to human reasoning has been argued to be
necessary for the success of artificial intelligence systems
\citep[\eg,][]{levesque.etal.2011}. One well-studied domain that requires
reasoning is question answering, where simply memorizing and looking up
information is often not enough to correctly answer a question.
For example, given the very simple scenario in \Table{tab:babi}, searching for
the word ``Mary'' and returning a nearby word is not a correct strategy;
instead, a model needs to recognize that Mary is currently at the second
location (office and not the bathroom).

Recent research has focused on developing neural models that succeed in such
scenarios \citep{sukhbaatar2015end,henaff2016tracking}. As a benchmark to
evaluate these models, \citet{weston2016towards} released a dataset -- Facebook
bAbi -- that provides a set of toy tasks, each examining a specific type of
reasoning. For example, the scenario in \Table{tab:babi} evaluates the capacity
to reason using a single supporting fact. However, the bAbi tasks are already
too simple for the current models. Only a few years after their release,
existing models fail at only one or two (out of 20) tasks
\citep{rae.etal.2016,santoro.etal.2017}. Moreover, all except two of the
reasoning tasks in this dataset only require transitive inference
\citep{Lee2015ReasoningIV}.
\begin{table}[h]
\vspace{-0.5cm}
\footnotesize
\centering
\begin{tabular}{|l|}
\hline
Mary went to the bathroom. \\
John moved to the hallway. \\
Mary travelled to the office.\\
Where is Mary? A: office\\
\hline
\end{tabular}
\vspace{-0.3cm}
\caption{\small{A task from the bAbi dataset \citep{weston2016towards}.}}
\label{tab:babi}
\vspace{-0.5cm}
\end{table}

People reason not just about their own observations and beliefs but also about
others' mental states (such as beliefs and intentions). The capacity to
recognize that others can have mental states different than one's own --
\textit{theory of mind} -- marks an important milestone in the development of
children and has been extensively studied by psychologists \citep[for a review,
see][]{flavell.2004}.   
Artificial intelligence (AI) systems will also require a similar reasoning
capacity about mental states as they are expected to be able to interact with
people \citep[\eg,][]{chandrasekaran2017takes,
grant.etal.2017,rabinowitz.etal.2018}.

However, the bAbi dataset does not include tasks that evaluate a model's
ability to reason about beliefs.  \citet{grant.etal.2017} created a bAbi-style
dataset inspired by an influential experiment on the theory of mind called the
Sally-Anne task \citep[\eg][]{baron1985does}. Their goal was to examine whether
the end-to-end memory network \citep{sukhbaatar2015end} can answer questions
such as ``where does Sally think the milk is?'' in situations that Sally's
belief about the location of milk does not match the reality. For example,
Sally thinks that the milk is in the fridge but the milk is actually on the
table.

The dataset of \citet{grant.etal.2017} provides a first step in designing
benchmarks to evaluate the mental-state reasoning capacity of
question-answering models, but it is still limited in the types of reasoning it
probes. For example, it only considered first-order beliefs (\eg, Sally's
belief about the location of milk). People also reason about second-order (and
higher-order) beliefs (\eg, Anne's belief about Sally's belief about the
location of the milk). 
More importantly, similarly to the bAbi dataset, success in each task is
defined as correctly answering one question. This does not guarantee that a
model has an understanding of the state of the world; in fact, even in
developmental theory-of-mind experiments, children are asked a few questions
(\eg, ``where is milk really?'') to ensure that their correct answer reflects
their understanding and is not simply due to chance. 

In this paper, we address these shortcomings by designing a new dataset that
enables us to evaluate a model's capacity to reason about different types of
beliefs as well as whether it maintains a correct understanding of the world.
To this end, we evaluate a number of different models that perform well on the
bAbi tasks: the end-to-end memory network \citep{sukhbaatar2015end}, the
multiple observer model \citep{grant.etal.2017}, the recurrent entity network
\citep{henaff2016tracking}, and RelationNetwork \citep{santoro.etal.2017}.  We
find that none of these models succeed at our tasks, suggesting that they are
not able to keep track of inconsistent states of the world, in particular when
someone's belief does not match the history or reality of a situation.

\section{Theory of Mind Experiments}
\label{sec:data}

Behavioral research shows that children gradually develop a theory of mind
\citep[for a review, see][]{gopnik1988children}. At the age of two, most
children have an understanding of others' desires and perceptions -- if someone
wants something, they will try to get it and if something  is in their sight,
they can see it. Children begin to understand others' beliefs around the age of
three, but this understanding is still limited. For example, they might not be
able to reason that someone's actions are a result of their beliefs. By the age
of five, most children have a unified theory of mind and are able to represent
and reason about others' desires, perceptions, and beliefs. Developmental
psychologists have designed various experimental paradigms to examine to what
extent children are able to reason about others' mental states. We use these
experiments as guidelines for designing tasks to evaluate the reasoning
capacity of question-answering models. We first explain these experiments.

\subsection{The Sally-Anne Experiment}

\begin{figure}[h]
\centering
\includegraphics[scale=0.4]{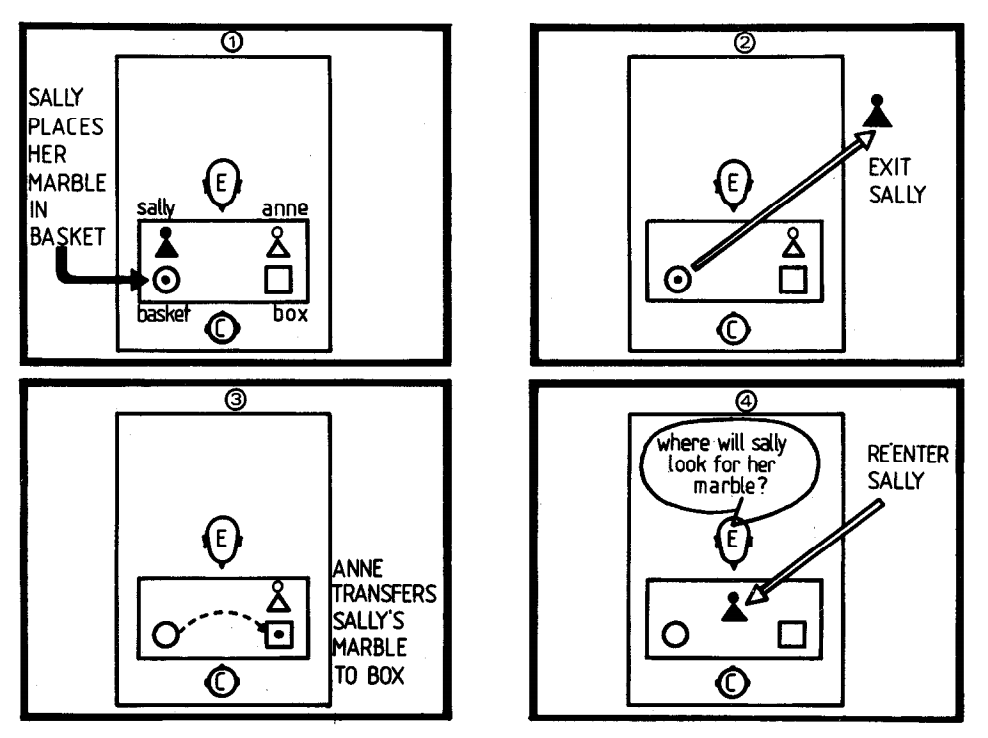}
\caption{\small The Sally-Anne experiment setup from \citet{baron1985does}.} 
\label{fig:sally-anne-exp}
\vspace{-0.5cm}
\end{figure}

The Sally-Anne false-belief experiment, proposed by \citet{baron1985does},
examines children's ability to reason about others' \textit{false beliefs},
\ie, when someone's belief does not match the reality. In this experiment, the
participants observe two agents, Sally and Anne, with their containers, a
basket and a box. After putting a marble in her basket, Sally leaves the room
(and is not able to observe the events anymore). After Sally's departure, Anne
moves the marble to her box. Then, Sally returns to the room (see
\Figure{fig:sally-anne-exp}). The participants are asked the following
questions:

\begin{itemize}
    \item ``Where will Sally look for her marble?''   \\ \textit{(belief question)}  \vspace{-0.2cm}
    \item ``Where is the marble really?''             \\ \textit{(reality question)} \vspace{-0.2cm}
    \item ``Where was the marble in the beginning?''  \\ \textit{(memory question)}  \vspace{-0.2cm}
\end{itemize}

The first question tests the participants' ability to reason about Sally's
belief about the location of her marble. Interestingly, most children before
the age of 3 answer this question incorrectly and say that Sally will look at
the box (where the marble really is) instead of the basket (where Sally thinks
the marble is). These children are not able to reason about Sally's belief
which is different from the reality of the world. 
The reality and memory questions are used to confirm that children's correct
answer to the belief question is not due to chance; but because they have a
correct understanding of the state of world and others' beliefs. 

\subsection{The Icecream Van Experiment}
The Sally-Anne experiment examines the ability to reason about another person's
belief or a \textit{first-order belief}. People are also able to reason about
beliefs about beliefs, for example, Anne thinks that Sally believes that the
marble is in basket. \citet{perner.wimmer.1985} performed a set of experiments
to examine children's reasoning capacity about such higher-order beliefs. In
their set-up Mary and John together see an ice cream van in the park, and the
icecream man tells them that he will be in the park until later in the
afternoon. Mary leaves the park and goes home. A bit after she leaves, the
icecream man decides to leave the park and tells John that he is going to the
church. On his way to the church he runs into Mary and informs her that he will
be selling icecreams close to the church all afternoon. The participants are
then asked the following second-order question: ``Where does John think Mary
goes to get icecream?'' Note that John does not know that Mary has been told
about the new location of the icecream van; he has a \textit{second-order false
belief} about Mary's belief. The participants are also asked a few control
questions (\eg, ``does Mary know that the van is in the church?'') to ensure
that they do not correctly answer the second-order question by chance.
\citet{perner.wimmer.1985} found that  6- and 7-year old children are able to
answer the second-order questions, suggesting that reasoning about higher-order
beliefs (as compared to a first-order belief) is a harder cognitive task.

\section{The Theory of Mind Task Dataset}

Inspired by the theory-of-mind experiments explained in \Section{sec:data} and
building on the work of \citet{grant.etal.2017}, we created a dataset based on
three tasks designed to capture increasingly complex theory-of-mind reasoning:
\emph{true-}, \emph{false-}, and \emph{second-order false-belief} tasks.
Examples of each task type are given in \Figure{fig:tombabi-data}.  In the
true-belief task, Sally observes the world and as a result she has a
first-order \emph{true belief} about the location of the milk -- her belief
matches reality.  In the false-belief task, Sally's first-order belief differs
from reality (\ie, she has a \emph{false belief}) because she was absent when
the state of the world changed.
In the second-order false-belief task, Sally observes the new location of the
milk; thus, she has a \emph{true belief} about the milk's location.  However,
Anne's belief about Sally's mental state does not match reality because Anne
does not know that Sally has observed the change in the environment.  As a
result, Anne has a \emph{false belief} about Sally's beliefs.
These tasks are more challenging than the bAbI scenarios, because a model needs
to learn whether each agent has a true or false belief about a given world
state to succeed, where the world state now includes the mental states of each
agent.

Note that we assume all containers are transparent in the underlying world;
whenever an agent enters a location, they become aware of the object’s true
location. We made this decision to keep the tasks as structurally similar as
possible. This  prevents models from simply learning to produce a specific
answer for a task type when a sentence like ``Sally looks inside the pantry''
is present in the story.  The container-transparency property is consistent
throughout all task-question pairs.

\begin{figure*}[!bht]{}
\resizebox{1 \textwidth}{!}{

\begin{tabular}{lll}
\textbf{True Belief} &
\textbf{False Belief} &
\textbf{Second-order False Belief} \\

 Anne entered the kitchen. &
 Anne entered the kitchen. &
 Anne entered the kitchen. \\
 
 Sally entered the kitchen. &
 Sally entered the kitchen. &
 Sally entered the kitchen. \\ 
 
 The milk is in the fridge. &
 The milk is in the fridge. &
 The milk is in the fridge. \\
 
 Anne moved the milk to the pantry. &
 \textit{Sally exited the kitchen.} &
 \textit{Sally exited the kitchen.} \\
 
 &
 Anne moved the milk to the pantry. &
 Anne moved the milk to the pantry. \\

 &
 &
 Anne exited the kitchen. \\

 &
 &
 \textit{Sally entered the kitchen.} \\
\end{tabular}
}
\vspace{-0.2cm}
\caption{\small An example story from each of the three task types.}
\label{fig:tombabi-data}
\vspace{-0.5cm}
\end{figure*}

\paragraph{Question types.} To examine the reasoning capacity of each model
about beliefs and second-order beliefs, we employ four question types inspired
by theory-of-mind experiments discussed in Section \ref{sec:data}; see
\Table{tab:tombabi-qs} for examples of these question types. These questions
enable us to test whether a model can reason about first-order and second-order
beliefs, and at the same time, knows the initial and current correct location
of an object; thus, we can distinguish between when a model answers a question
by chance and when it actually understands the entire state of the world.

\Table{tab:tombabi-tqpairs} gives the answers for the $12$ combinations of task
type and question. Given a \tb or \fb task, the answers to the first-order and
second-order questions are the same (\eg, ``pantry'' in the \tb condition and
``fridge'' in the \fb condition for the tasks in \Figure{fig:tombabi-data}).
However, they are different in the \textit{second-order false belief} task
because Anne has a false belief about Sally's belief.

\paragraph{Dataset variants.}
We use these tasks to generate two  datasets: \tom and \tomeasy. The primary
difference between these two datasets is that, in \tomeasy, each story has only
one task, while \tom can have multiple tasks within a single story. Each
dataset contains a training set with 10 000 examples with each of the 12
combinations of task and question types. 

In \tom, the tasks are randomly grouped into sets of 5 to form stories, which
is the same number used in the bAbI dataset. In the test set for \tom, each
story contains 4 tasks, but there is only one question present at the end.
Because questions that come closer to the beginning of a story have fewer
distracting sentences (\ie, potential answer words) that may confound a model,
they are easier to answer.  We found that this testing procedure gave us a more
precise understanding of the performance of the model by separating the
difficulty of a question due to its position in a story from the inherent
difficulty of the question itself.

\paragraph{Generating the data.}
Each reasoning task in \citet{weston2016towards} can be formalized with a
grammar. The training and test data are then the derivations of this grammar.
We refer to each derivation as a story (\eg, \Figure{fig:tombabi-data}). We
follow \citet{grant.etal.2017} in writing grammars for our new tasks. 
In particular, all the task grammars consist of a set of \textit{entities}
(people and objects in the stories) and \textit{predicates} that take entities
as subject or object. The grammars also specify the \textit{properties} of
entities -- which predicates take them as subjects or objects. A predicate can
include \textit{actions} that are ways an agent interact with the world (\eg,
\textit{place, move, enter, exit}) and \textit{beliefs} that are mental state
terms (\eg, \textit{believe, think}).  As an example, \textit{Sally} with the
property \textit{is agent} can perform the action \textit{displace} on
\textit{apple} with the property \textit{is object}.  Similar to the previous
work, we use a restricted set of \textit{action} and \textit{belief}
predicates.

\begin{table}[t]
    \centering
    \begin{tabular}{l|l}
         \small{Memory} & \scriptsize{Where was the milk at the beginning? } \\
         \small{Reality} &  \scriptsize{Where is the milk really?} \\
         \small{First-order}  & \scriptsize{Where will Sally look for the milk?}\\
         \small{Second-order}& \scriptsize{Where does Anne think that Sally searches for the milk?}
    \end{tabular}
    \vspace{-0.2cm}
    \caption{\small Examples of the four question types.}
    \label{tab:tombabi-qs}
    \vspace{-0.2cm}
\end{table}

\begin{table}
    \begin{tabular}{l|lll}
                & \textbf{TB} & \textbf{FB} & \textbf{SOFB} \\ \hline
        Memory       & first  & first  & first \\
        Reality      & second & second & second \\
        First-order  & second & first  & second \\
        Second-order & second & first  & first 
    \end{tabular}
    \vspace{-0.2cm}
    \caption{\small The correct answer to each question for \tb (TB), \fb (FB), and \sofb (SOFB) tasks. Here, ``first'' and ``second'' are the initial and actual locations of the object of interest, respectively (\eg, fridge and pantry in  \Figure{fig:tombabi-data}).
    }
    \label{tab:tombabi-tqpairs}
    \vspace{-0.2cm}
\end{table}

\section{The Models}

We briefly describe the models that we evaluate in this paper. We chose these
models based on the novelty in their architecture or their near
state-of-the-art results in the bAbi tasks. 
More specifically, given 10k examples and joint-training on the bAbi tasks, the
best end-to-end memory network \citep{sukhbaatar2015end} and relation network
\citep{santoro.etal.2017} fail at $6$ and $2$ tasks, respectively. Given the
same training dataset and per-task training, the recurrent entity network
succeed at all tasks (but the authors do not report the results of
joint-training).  Recall that the bAbi tasks are structured as a set of
sentences followed by a question about them (\eg, a story from
\Figure{fig:tombabi-data} followed by a question from \Table{tab:tombabi-qs}).

\noindent\textbf{The End-to-End Memory Network.}
\citet{sukhbaatar2015end} proposed a neural memory-augmented model, the
end-to-end memory network (MemN2N), that extends the memory network
architecture \citep{weston.2014arXiv}. Similarly to its predecessor, MemN2N has
a memory component in which sentences are embedded and an attention function
that weights the embedded sentences based on their similarity to a given
question. The MemN2N model introduces multiple layers of memory (hops) by
stacking memory components such that the question embedding at layer $k+1$ is
the sum of the output and question embedding of layer $k$. 

\noindent\textbf{The Multiple Observer Model.}
To perform well on the \fb and \sofb conditions, a model needs to identify that
agents have experienced different events, and, as a result, have differing
knowledge about the state of the world.  Although having multiple layers of
memory in the MemN2N model enables it to combine attention to different memory
slots (\ie, embedded sentences) at each layer, the model does not have access
to each agent's unique perspective. For example, the model is not explicitly
told that Sally does not observe the change of the location of the milk in the
\fb condition.
To address this, \citet{grant.etal.2017} propose the Multiple Observer model
that integrates MemN2N with individual memory modules for each agent in the
story. An agent's memory only receives the sentences for which the agent is
present and observes the world. Their model has an additional attention
function that weighs the memory modules based on their relevance to the
question. The model is expected to learn to attend to Sally's memory module if
the question is about her belief about a state of the world.

\noindent\textbf{The Recurrent Entity Network.}
\citet{henaff2016tracking} propose a memory-augmented architecture, EntNet,
with two interesting properties; first, their model is a recurrent neural
network and thus can capture the sequential nature of the events in a story.
Second, instead of keeping a whole sentence embedding in a memory slot, their
model can learn the important entities of a story (\eg, a person) and their
properties (\eg, location) through a set of gated recurrent units and two
weight matrices. 

\noindent\textbf{The Relation Network.} \citet{santoro.etal.2017} propose a
neural model for relational reasoning. Their model consider the possibility of
a relation among each two possible pairs of objects. To model the bAbi tasks,
they consider each pair of sentences together with the question as inputs to
their relation network.

\section{Experimental Results}

\paragraph{Experiment set-up}

We train all models jointly over all task types without noise, but evaluate
them independently on different task and question pairs. We choose the
best-performing models by selecting hyperparameters on the validation set.
{Similarly to \citet{sukhbaatar2015end}, we consider a model successful only
when its accuracy exceeds $95\%$ across the entire task suite.} 

\paragraph{MemN2N and Multiple Observer Models.} We first examine how each
model performs across a range of parameter and initialization values. MemN2N
models are very sensitive to the network initialization and for each set of
parameters, the best result out of 10 runs is reported
\citep{sukhbaatar2015end}. We first visualize the accuracy of all runs as a box
plot to identify how sensitive each model is to random initialization (of
parameters and internal states) and thus difficult to train. We also report the
results for the best run in each experiment. We use a memory size of 50, the
same as experiments of \citet{sukhbaatar2015end}, to ensure that the memory
contains all sentences of a given story.

\paragraph{EntNet.} We report results averaged over 3 initializations because
we observed little randomness due to initialization. We selected the learning
rate on a held out validation set separately for \tomeasy and \tom; all otehr
the same hyperparameters as \citet{henaff2016tracking}: 20 memory slots and an
embedding size of 100 We trained until the training error hit zero, which
occurred around 50 epochs for both datasets. . 

\paragraph{RelNet.}
We report results using a single seed because we saw little randomness due to
initialization; this is in accordance with the authors'
findings~\citep{santoro.etal.2017}.
We selected model hyperparameters on a held-out validation set separately for
each of the \tom and \tomeasy datasets.

\begin{figure*}[!bht]
         \centering
	 \caption{Memory Network and Multiple Observer Model Performance Across
Task and Question Types. \small{Pink indicates that the answer to the question
is the first container that contained the object in that task. Blue indicates
that the answer is the last container that contained the object before the
question was asked. Grey indicates that the answer was the first container that
contained the object in the entire story which may or may not be the same as
the pink.}} \vspace{-.2cm}
         \begin{subfigure}{\textwidth}
            \caption{Memory Network with memory size 50 evaluated on the \tomeasy dataset.}
            \vspace{-0.2cm}
            \centering
            \includegraphics[width=\linewidth]{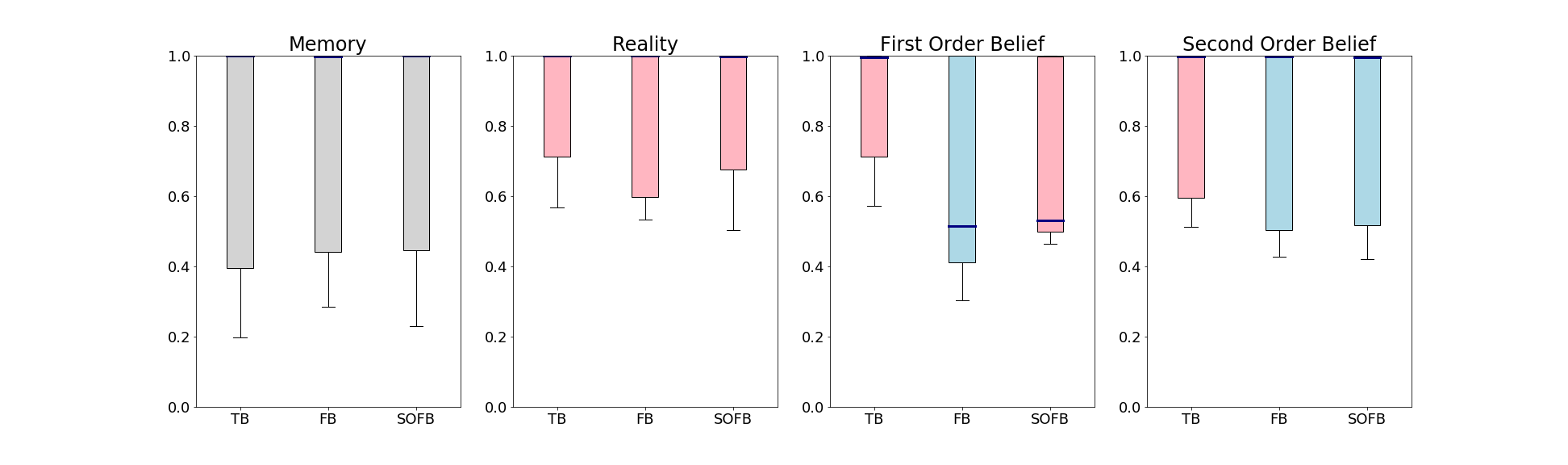}
            \label{fig:1c1t1t10k50m0noise}
            \vspace{-0.2cm}
        \end{subfigure}
        \begin{subfigure}{\textwidth}
            \caption{Multiple Observer Model with memory size 50 evaluated on the \tomeasy dataset.}
            \centering
            \vspace{-0.2cm}
            \includegraphics[width=\linewidth]{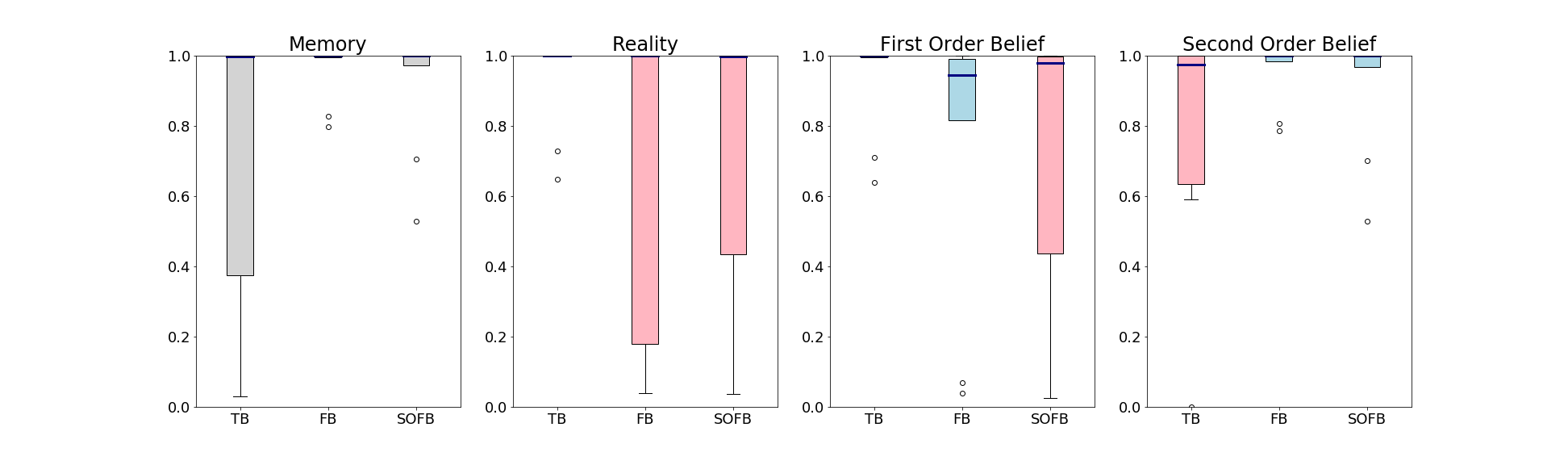}
            \label{fig:5c1t1t10k50m0noise}
            \vspace{-0.2cm}
        \end{subfigure}
        \begin{subfigure}{\textwidth}
            \caption{Memory Network with memory size 50 evaluated on the \tom dataset.}
            \vspace{-0.2cm}
            \centering
            \includegraphics[width=\linewidth]{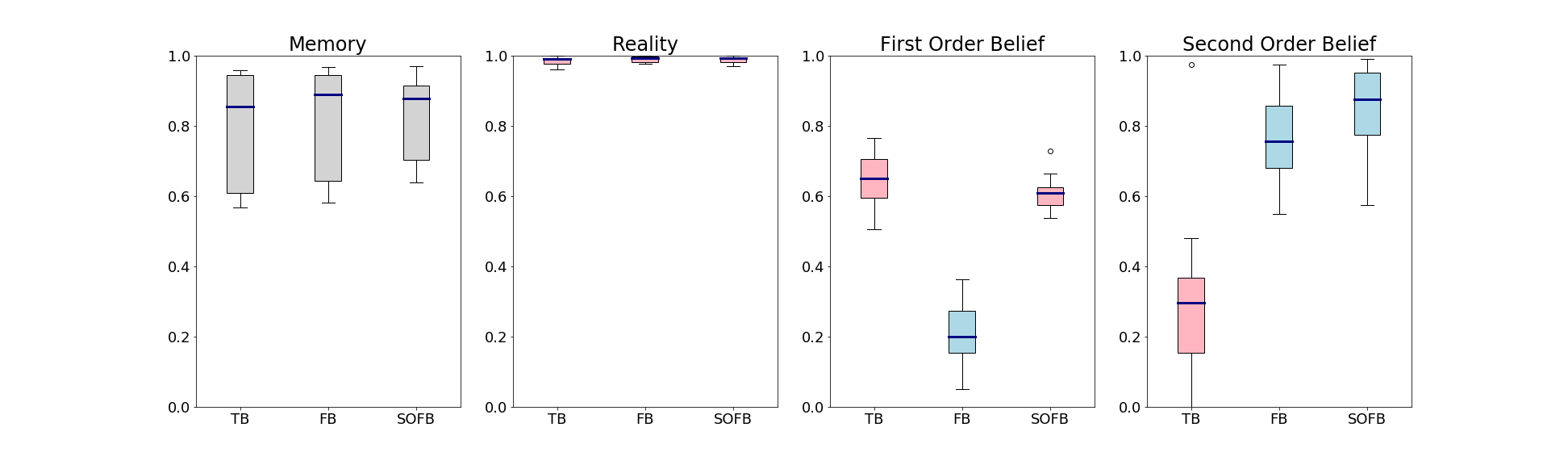}
            \label{fig:1cNtNt10k50m0noise}
            \vspace{-0.2cm}
        \end{subfigure}
        \begin{subfigure}{\textwidth}
            \caption{Multiple Observer Model with memory size 50 evaluated on the \tom dataset.}
            \vspace{-0.2cm}
            \centering
            \includegraphics[width=\linewidth]{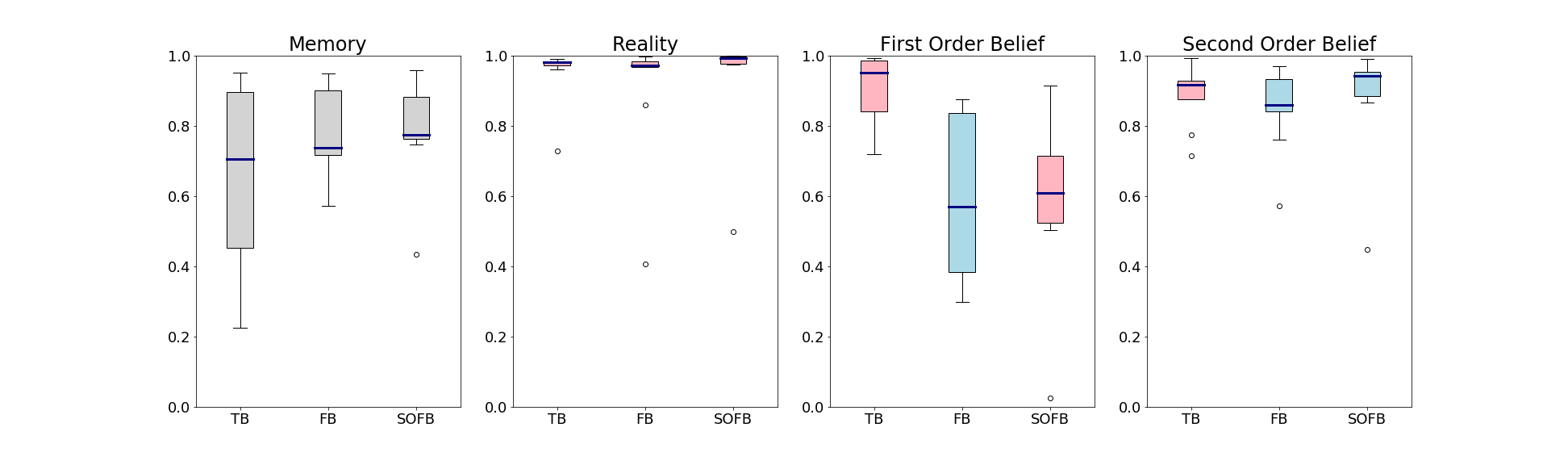}
            \label{fig:5cNtNt10k50m0noise}
            \vspace{-0.2cm}
        \end{subfigure}
       \vspace{-0.7cm}
 \end{figure*}

\subsection{Overall Performance on \tomeasy}

We expect the models perform well on this dataset, given that there is only one
task in the memory during both training and test and as a result unrelated
events do not interfere with a model's reasoning in this condition.  

Despite the large variance in accuracy across runs, the MemN2N models often
succeed at the memory, reality, and second-order questions
(\Figure{fig:1c1t1t10k50m0noise}). Note that the median accuracy (the dark blue
line in box) is close to 1 for these questions. However, the model often fails
(median accuracy around 0.5) given the first-order question (``where will Sally
look for the milk'') and  the \fb and \sofb tasks. This pattern is different
from the empirical findings on people; the second-order question is harder than
the first order one for people. We observe that the performance of the Multiple
Observer model is similar to the MemNet for most question-task pairs (expect
one) but there is less variation in the accuracy values. Interestingly, the
median accuracy is close to one for the first-order question and the \sofb task
but the Multiple Observer model still performs poorly for this question on the
\fb task.

Why is the first-order question harder for the MemN2N model? To investigate
this, we look more closely at our task-question pairs. As shown in
\Table{tab:tombabi-tqpairs}, the answers to the first-order question are
different for the \fb and \sofb tasks but are the same for the second-order
one. We suspect that it is harder for the MemN2N model to learn two distinct
answers for the same question given the the similarities of the two
false-belief tasks. To test this hypothesis, we altered the data such that the
answers to first-order question are (incorrectly) the same for both \fb and
\sofb tasks (and also the second-order question). We observe that the median
accuracy is close to 1 for all conditions suggesting that the model can learn
the distinction between two of the tasks but not all three.

We observe that EntNet and RelNet models are not too sensitive to the
initialization value, and thus just report the result on best-performing
models.  Both EntNet and RelNet best models succeed at the \tomeasy tasks;
their mean error is $0$.

\subsection{Overall Performance on \tom}

This dataset is more similar to the bAbi dataset in that, during both training
and test, the memory contains a story with multiple tasks; as a result, it is
harder for the model to identify the entities relevant to a given question.

{As shown in \Figure{fig:1cNtNt10k50m0noise}, the MemN2N performs worse on all
the questions with the exception of the reality question, where performance is
slightly worse but has lower variance.} (\cf the \tomeasy dataset). The first-
and second-order questions are, overall, harder for the model. The performance
of the Multiple Observer model is better for most of the questions (see
\Figure{fig:5cNtNt10k50m0noise}), especially the second-order, but it is
slightly worse for the memory question. This suggests that adding
agent-specific memory modules is not enough to succeed on all the \tom tasks,
and that the increased complexity of inference with multiple memory modules
harms performance on another task.

We also look at the best-performing MemN2N and Multiple Observer models over
all question-task pairs.  These models are selected based on their performance
on the validation dataset. We observe that none of these models succeed on the
\tom tasks, but, interestingly, the Multiple Observer model performs better
overall as compared to the original MemN2N model (see
\Table{tab:entnet-relnet-results}).

\paragraph{\tom and bAbi.} How are \tom and bAbi tasks similar? The combination
of our \tb task and the reality question is very similar to bAbi task 1 (single
supporting fact). To correctly answer the reality question (``where is the milk
really?'', a model need to use a single fact from the story (``Anne moved the
milk to the pantry.''). The MemN2N model succeeds at both bAbi task 1 and the
reality question given the \tb task.
However, the correct answer to the memory question (``where was the milk at the
beginning?'') for the \tb task also requires a single fact (``the milk is in
the fridge.''). Interestingly, the error of MemN2N on the memory question is
much higher than the reality question. The model (unlike people) cannot learn
two representations (initial and current location) for an object. This result
demonstrates the importance of representing alternative states of the world,
whether it be past states of reality (\ie, where the ``milk'' used to be) or
mental states about the world (\ie, where an agent believes the ``milk'' to
be).

\newcommand{\fillmein}{\textcolor{red}{????}}
\begin{table*}[t!]
\begin{center}
\small
\begin{tabular}{lcccc}
    \toprule
    Model  & \textbf{\tomeasy} & \textbf{\tomeasy} (noised) & \textbf{\tom} & \textbf{\tom} (noised) \\
    \midrule
    MemN2N~\citep{sukhbaatar2015end}          & $100.00\%$ & $90.28\%$ & $82.38\%$ & $77.00\%$ \\
    Multiple Observer~\citep{grant.etal.2017} & $100.00\%$ & $\mathbf{95.81}\%$ & $91.11\%$ & ${87.43}\%$ \\
    EntNet~\citep{henaff2016tracking}         & $100.00\%$ & $94.61\%$ & $93.67\%$ & $\mathbf{88.63}\%$ \\
    RelNet~\citep{santoro.etal.2017}          & $100.00\%$ & $87.82\%$ & $\mathbf{94.31\%}$ & $76.84\%$ \\ 
    \bottomrule
\end{tabular}
\vspace{-0.2cm}
\caption{\small 
    A comparison of model performance. All models succeed on the \tomeasy dataset without noise. Multiple Observer model performs best on \tomeasy with noise, RelNet performs best on \tom, and EntNet performs best on \tom with noise.
}
\label{tab:entnet-relnet-results}
\vspace{-0.5cm}
\end{center}
\end{table*}

\begin{table*}[t!]
\begin{center}
\scriptsize{
\begin{tabular}{l|c|c|c}
    \toprule
    Model  & \textbf{True-belief} & \textbf{False-belief}  & \textbf{Second-order False-belief} \\
    \midrule
    MemN2N~\cite{sukhbaatar2015end}          & M ($94.4$); FOB ($78.2$); SOB ($42.9$) & M ($94.3$); FOB ($17.3$)               & M ($92.8$); FOB ($56.4$) \\
    Multiple Observer~\cite{grant.etal.2017} & M ($93.2$)                             & M ($90.5$); FOB ($56.4$); SOB ($92.5$) & M ($93.0$); FOB ($90.3$); SOB ($90.3$) \\ 
    EntNet~\cite{henaff2016tracking}         & M ($74.0$)                             & M ($76.1$)                             & M ($74.25$)\\
    RelNet~\cite{santoro.etal.2017}          & M ($39.7\%$); FOB ($71.37\%$)          & M ($78.5\%$)                           & M ($34.9\%$); FOB ($81.8\%$) \\
    \bottomrule
\end{tabular}
}
\vspace{-0.2cm}
\caption{\small 
    Model accuracy on failed questions given the \tom task (without noise); M, R, FOB, and SOB are the memory, reality, first- and second-order questions, respectively. 
    The number in the parentheses is the accuracy for that question on that task.
}
\label{tab:failedq}
\vspace{-0.5cm}
\end{center}
\end{table*}

\paragraph{EntNet and RelNet.} We also report results of two relevant
memory-augmented neural network models on our tasks,
EntNet~\citep{henaff2016tracking} and RelNet~\citep{santoro.etal.2017}, in
\Table{tab:entnet-relnet-results}.  Again, because we did not observe
sensitivity to initialization for these models, only average performance of
their best-performing model is reported.
We see that even though these models succeed on the \tomeasy dataset, they fail
on the \tom tasks, suggesting that these models cannot simultaneously deal with
inconsistent (\ie, past, present, and mental) states of the world.

We further investigate which questions are the hardest for each best model; see
\Table{tab:failedq}. We observe that each of the MemN2N, Multiple Observer and
RelNet models perform poorly on some combination of the first- and second-order
questions, but are successful at answering the reality question.  We
hypothesize that this phenomenon occurs because the reality question is the
most similar to the bAbi tasks.
In addition, all models fail the memory question for each task type.  While
this is to be expected for EntNet due to its recurrent nature and therefore
bias towards recency, it is surprising that the other models, which exhibit
only a small positional bias, cannot correctly represent a past state of the
world in order to answer the memory question correctly.

\subsection{Experimenting with Noise} We examine to what extent each model's
architecture is sensitive to the position of sentences in each story. We do so
by adding a novel sentence at random locations in each story at test time. For
any setting of the noise, $p$, there is a $p\%$ probability of a noise sentence
occurring before each sentence in the story. Noise sentences cannot follow
other noise sentences in the story. {In this paper, we report results with
$p=.1$.}
We observe that the accuracy of all best models decreases notably in the
presence of noise (see \Table{tab:entnet-relnet-results}). This result is
particularly interesting as it shows that none of the models are able to use
the semantics of the sentences in a story in their reasoning -- they are all
sensitive to the presence of distractor sentences.

Interestingly, the RelNet model is the best performer amongst the models we
considered on the \tom dataset, yet it is also the most sensitive to noise.
Moreover, the Multiple Observer model -- with explicit memories for each agent
-- is the most robust to noise; it has the minimum decrease in accuracy between
each dataset and its noised version.

\subsection{Experimenting with Memory} 
In the experiments of \citet{sukhbaatar2015end} the memory size is fixed to 50,
which is necessary to capture the entire story in memory (e.g. the answer to
the memory question in \tom may rely on information at the beginning of a
story).
We observed that smaller memory sizes artificially improved the performance of
the MemN2N and Multiple Observer model on \tom tasks. For example, using a
memory size of 10, our best MemN2N model performance boosts on the hardest task
of \tom (FB task with first order belief question) from $5.1\%$ to $97.5\%$ and
on the easiest task from $98.3\%$ to $100.0\%$ (SOFB task with reality
question).
This result is not surprising because given a small memory size, \tom and
\tomeasy are very similar tasks; the memory size of 10 allows for at most two
full tasks in memory.

\section{Related Work}

Recent research has emphasized the importance of modeling and understanding
people's mental states for AI systems. \citet{eysenbach2016mistaken} created a
dataset of scene-description pairs where each scene is a set of visual frames
and some frames include people with mistaken beliefs.\footnote{For example, a
scene where a person gets sick eating mushrooms is paired with the sentence
``the couple mistakenly thinks it's ok to eat the mushrooms''.}
The authors build a regression model for identifying a mistaken belief and the
person who has such a belief in a given frame. Our work differs with theirs in
that we are interested in understanding whether a model can reason about
people's true and false beliefs to correctly answer questions as opposed to
identifying mistaken beliefs. 

\citet{grant.etal.2017} studied whether the end-to-end memory network of
\citet{sukhbaatar2015end} can pass a false-belief test -- correctly answer
where Sally would search for an object in false- and true-belief situations.
They created a dataset inspired by the bAbi dataset to examine whether the
model can reason about interaction of beliefs and actions in these situations
-- how actions cause beliefs and vice versa.  They show that MemN2N fails at
the false-belief test, and their extension of that model with separate memories
for each agent and an observer outperforms MemN2N.

\citet{rabinowitz.etal.2018} formulate the capacity to reason about others'
beliefs as a meta-learning problem. They propose a neural network, ToMnet, that
learns to predict the behavior of different agents given their past and current
trajectory. Similarly to \citet{grant.etal.2017}, in addition to individual
agents, they model an ``observer'' that has access to states and actions of all
agents (though this information can be noisy and partial). Interestingly, their
model successfully predicts an agent's behavior in a false-belief situation --
the agent's behavior reflects its false-belief as opposed to the reality of the
world.

Finally, \citet{chandrasekaran2017takes} take a different approach by studying
whether people can understand the ``beliefs'' of a visual-question answering
system. More specifically, they examine whether the participants can predict
when the model would fail in answering a question as well as if they can
predict the model's answer. They find that even with a few examples, people get
better at answering these questions.

\section{Discussion}
We propose a dataset for evaluating question-answering models. Our dataset --
inspired by seminal theory-of-mind experiments in children -- measures to what
extent recently introduced neural models can reason about beliefs and states of
the world that are potentially mututally inconsistent.
We evaluate three of the recent neural question answering models (and an
extension of one) on our tasks. We find that none of the models are able to
succeed fully on a suite of tasks that requires keeping track of inconsistent
beliefs or states of the world.
These inconsistencies arise from differences between the past and the present,
as well as the mental states of agents who may have false beliefs about the
world or about the mental states of other agents.

The purpose of the dataset introduced in this work is not to test advanced
language fluency; instead, consistency in the linguistic structure of the tasks
allows us to isolate the performance of the models' reasoning capabilities.
Even though the language is simple, the models struggle to achieve good
performance.
Furthermore, we note that the proposed dataset should be treated as a
diagnostic tool and that good performance on similar toy tasks is not
sufficient for reasoning capabilities.

\bibliography{references}
\bibliographystyle{acl_natbib_nourl}

\end{document}